\documentclass[journal]{IEEEtran}
\usepackage{graphicx}  
\usepackage{algorithm}
\usepackage{booktabs} 
\usepackage{algorithmic}
\usepackage{graphicx}
\usepackage{multirow}

\usepackage{times}
\usepackage{epsfig}
\usepackage{graphicx}
\usepackage{amsmath}
\usepackage{amssymb}
\usepackage{graphicx}  
\usepackage{algorithm}
\usepackage{booktabs} 
\usepackage{algorithmic}
\usepackage{graphicx}
\usepackage{multirow}
\usepackage{color}

\begin{document}
%
\title{PDNet: Prior-Model Guided Depth-enhanced Network for Salient Object Detection}
%
%
%

\author{Chunbiao~Zhu,~\IEEEmembership{Student~Member,~IEEE,}
        Xing~Cai,~\IEEEmembership{Student~Member,~IEEE,}
        Kan~Huang,~\IEEEmembership{Member,~IEEE,}
        Thomas.H~Li,~\IEEEmembership{Member,~IEEE,}
        and~Ge~Li,~\IEEEmembership{Member,~IEEE,}
\thanks{The authors are with the School of Electronic and Computer Engineering,
Peking University Shenzhen Graduate School, Shenzhen 518055, China. Prof.Ge Li(geli@ece.pku.edu.cn) is the corresponding author. This work was supported by the grant of National Natural Science Foundation of China(No.U1611461), the grant of Science and Technology Planning Project of Guangdong Province, China(No.2014B090910001) and the grant of Shenzhen Peacock Plan(No.20130408-183003656).
}
}

%
%

\markboth{Under review}%
{Zhu \MakeLowercase{\textit{et al.}}: PDNet}
%



\maketitle

\begin{abstract}
Fully convolutional neural networks (FCNs) have shown outstanding performance in many computer vision tasks including salient object detection. However, there still remains two issues needed to be addressed in deep learning based saliency detection. One is the lack of tremendous amount of annotated data to train a network. The other is the lack of robustness for extracting salient objects in images containing complex scenes. In this paper, we present a new architecture$ - $PDNet, a robust prior-model guided depth-enhanced network for RGB-D salient object detection. In contrast to existing works, in which RGB-D values of image pixels are fed directly to a network, the proposed architecture is composed of a master network for processing RGB values, and a sub-network making full use of depth cues and incorporate depth-based features into the master network. To overcome the limited size of the labeled RGB-D dataset for training, we employ a large conventional RGB dataset to pre-train the master network, which proves to contribute largely to the final accuracy. Extensive evaluations over five benchmark datasets demonstrate that our proposed method performs favorably against the state-of-the-art approaches.
\end{abstract}
\section{Introduction}
\label{sec:intro}
When human look at an image, he/she always focus on a subset of the whole image, which is called visual attention. Visual attention is a neurobiological process to filter out irrelevant information and highlight most noticeable foreground information. A variety of computational models have been developed to simulate this kind of mechanism, which can be used in object tracking~\cite{8265388}, image montage~\cite{Zhu2018} and image compression~\cite{Zhu2018An}. In general, saliency detection algorithms can be categorized into two groups: top-down~\cite{lee2016deep, Han2017CNNs,8265548} or bottom-up~\cite{crm2016spl,qin2015saliency, Shi2016, Li2015,Zhu2017} approaches. Top-down approaches are task-driven and need supervised learning. While bottom-up approaches usually use low-level cues, such as color features, distance features and heuristic saliency features. One of the most used heuristic saliency features is contrast, such as pixel-based or patch-based contrast.

Most previous works on saliency detection focus on 2D images. To our thoughts, this remains limited potential for further research. First, 3D data instead of 2D is more suitable for real application, second, as visual scene become more and more complex, utilizing 2D data only is not enough for extracting salient objects.
Recent advances in 3D data acquisition techniques, such as Time-of-Flight sensors and the Microsoft Kinect, have motivated the adoption of structural features, improving the discrimination between different objects with
the similar appearance. Saliency detection on RGB-D images will expedite a variety of real applications, such as 3D content surveillance, retrieval, and image recognition.

In addition to RGB information, depth has shown to be a practical cue for saliency estimation~\cite{8265548,8265388,Zhu2018}. However, it is still ineffective to train a network due to the limited size of annotated RGB-D data. Besides, how to integrate the additional depth information into the RGB framework remains to be a key issue that is needed to be addressed.

To resolve the above-mentioned limitation, in this paper, we propose a novel prior-model guided depth-enhanced network (PDNet). The PDNet is composed of a master network and sub-network. The master network is a convolution-deconvolution pipeline.
The convolution stage serves as a feature extractor that transforms the input image into hierarchical rich feature representation, while the deconvolution stage serves as a shape restorer to recover the resolution and segment the salient object in fine detail from background. The sub-network can be treated like an encoder convolution architecture and it process depth map as input and enhance the robustness of the master network. To address the problem of insufficient RGB-D data for training, we employ a large dataset to pre-train our master network. This pre-train setup before training our network using RGB-D data has proved to contribute dramatically to accuracy improvement. Fig.1 illustrates the pipeline of our model.

In summary, the main contributions of this work are as follows:
\begin{itemize}
  \item We propose a novel deep network (PDNet) for saliency detection on RGB-D images, where we utilize RGB-based prior-model to guide the main learning stage.
  \item Unlike the existing works, we process the depth cue in an independent encoder network, which can make full use of depth cues and assist the main-stream network.
  \item Compared with previous works, the proposed method demonstrates dramatical performance improvements on five benchmark datasets.
\end{itemize}
\begin{figure*}
\label{fig:1}
\begin{center}
\includegraphics[width=0.97\textwidth]{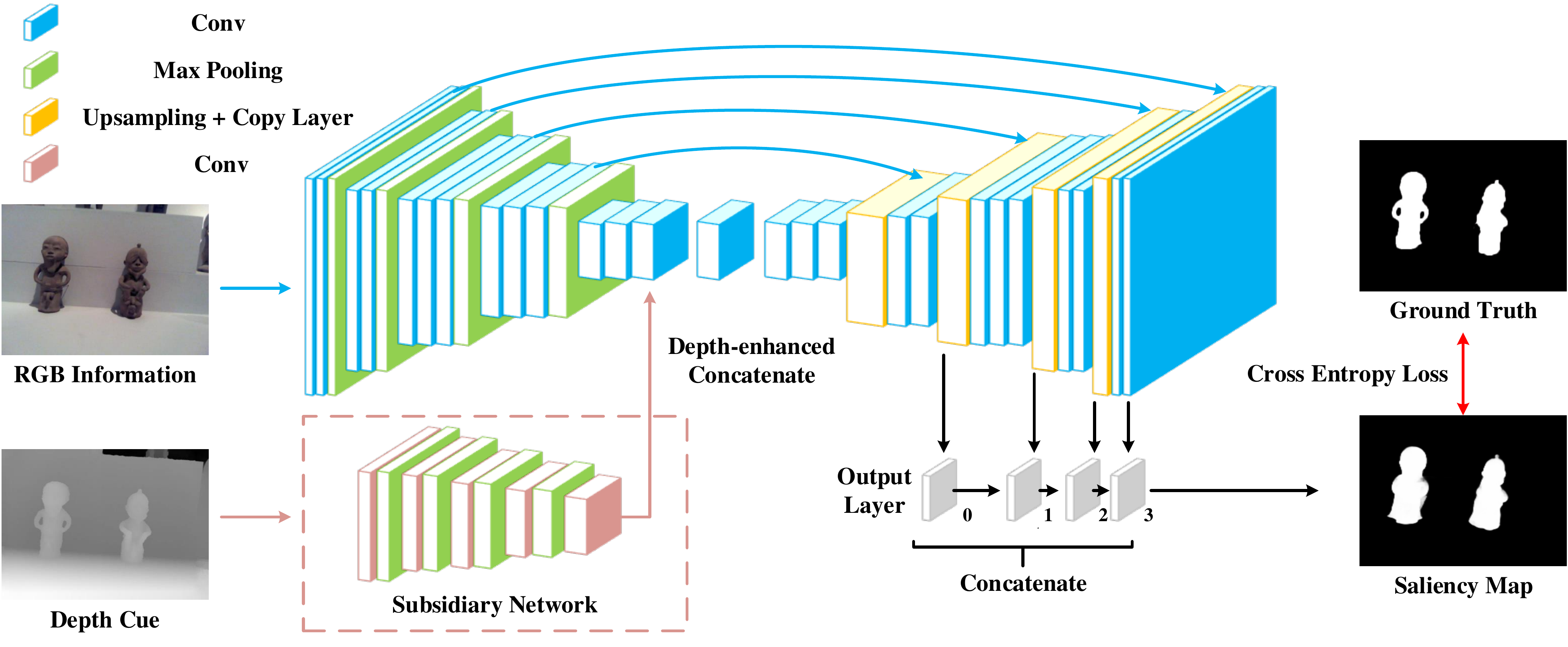}
\end{center}
\caption{The framework of the PDNet.}
\label{fig:short}
\end{figure*}
\section{Related Work}

In this section, we present a brief review of saliency detection methods on both RGB and RGB-D images.

\subsection{RGB Saliency Detection}
Over the past decades, lots of salient object detection methods have been developed. The majority of these methods are designed on low-level hand-crafted features~\cite{qin2015saliency, Shi2016, Li2015}. A complete survey of these methods is beyond the scope of this paper and we refer the readers to a recent survey paper~\cite{Borji2014Salient} for details.

Recently, with the development of deep learning and the growth of annotated data in RGB-based salient object detection datasets. The convolutional neural network has a remarkable performance in salient object detection. A lot of research efforts have been made to develop various deep architectures for useful features that characterize salient objects or regions. For instance, zhu et al.~\cite{8265548} presents a two-channeled perceiving residual pyramid networks to generating high-resolution and high-quality results for saliency detection. Li et al.~\cite{lee2016deep} fine tune fully connected layers of mutiple CNNs to predict the saliency degree of each superpixel. These methods achieve good performances, however, images with complex background are still a challenging task. Therefore, additionally auxiliary feature should be exploited to assist saliency detection.

\subsection{RGB-D Saliency Detection}
Compared with RGB saliency detection, RGB-D saliency has
received less research attention. In~\cite{Zhu2018}, Zhu et al. propose a framework based on depth mining, and use multilayer backpropagation to exploit the depth cue. In~\cite{Cheng2014Depth}, Cheng et al. compute salient stimuli in both color and depth spaces. In~\cite{Peng2014RGBD}, Peng et al. provide a simple fusion framework that combines existing RGB-produced saliency with new depth-induced saliency. In~\cite{Ju2015Depth}, Ju et al. propose a saliency method applied on depth images, which is based on anisotropic center-surround difference. In~\cite{Guo2016Salient}, Guo et al. propose a salient object detection method for RGB-D images based on evolution strategy.

However, for the limitation of the size of RGB-D datasets,  some deep learning based methods~\cite{Li2017CNN} use many pre-extracted low-level hand-crafted features to fed the network. And almost all the methods~\cite{Li2017CNN,Han2017CNNs} integrate the depth cue with RGB information directly as a fourth dimensional input to train the network. By contrast, our proposed method adopt pre-trained RGB network as prior-model and learn depth cue independently, which remedies the existing methods' drawback.

\section{Proposed Method}

As shown in Fig.1, the proposed PDNet contains two main components: the prior-model guided master network and the depth-enhanced subsidiary network. The master network is based on the convolution-deconvolution architecture. The subsidiary network acts like an encoder, extracting depth cues.
The proposed model will be discussed in detail in the following sections.
\subsection{Prior-model Guided Master Network}
\subsubsection{Master Network Architecture}
The master network is based on encoder-decoder architecture. VGG~\cite{simonyan2014very} is used in the encoder part of the proposed
model, besides, we employ copy-crop and multi-feature concatenation technique. We utilize hierarchical features in an effective way.

Here are the details of the proposed FCN network. Each of the convolution layers is followed by a Batch Normalization (BN) layer for improving the speed of convergence. And then the Rectified Linear Unit (ReLU) activation function is used for adding non-linearity. Every kernel size is $3 \times 3$ as used in other deep networks. VGG-16 and VGG-19 are tested for the encoder model. The experiment results will be shown in next section.

Copy-crop technique is used here for adding more low-level features from the early stage for improving fine details of saliency map on up-sampling stage.

Multi-feature concatenation technique is mainly based on loss-fusion pattern. It is used here for reasonably combining both low-level and high-level features for accurate saliency detection and loss fusion. Those features in different blocks in decoder part through one convolution kernel with $3\times3$ size and linear activation function get pyramid outputs. They are concatenated to final convolutional layer which has one $3\times3$ size kernel. The sigmoid activation function applied to this layer. Then, the pixel-wise binary cross entropy between predict saliency map $S$ and the ground truth saliency mask $G$ is computed by:
\begin{equation}
loss =  { \sum _{i = 0} ^{W} \sum _{j = 0} ^{H}} \frac{(1-G_{ij})\cdot log(1-S_{ij}) - G_{ij}\cdot log(S_{ij})}{{ W\times H}},
\end{equation}
where $i, j$ are the pixel location in an image.
\subsubsection{Prior-model Guidance}

Given an input image $I$, the salient object detection network produces a saliency map $Sm$ from a set of weights $\theta$. The salient object detection is posed as a regression problem, and the saliency value $S$ of each pixel $(i, j)$ in $Sm$ can be described as:
\begin{equation}
Sm_{i,j}= p(S|R(I,i,j); \theta),
\end{equation}
where $R(I, i, j)$ corresponds to the receptive field of location $(i, j)$ in $Sm$. Once the network is trained, $ \theta$ is fixed and used to detect salient objects for any input images.

Considering the limitation of RGB-D datasets, we employ the RGB based saliency detection datasets for pre-training. We utilize MSRA10K dataset~\cite{Cheng2011Global} as well as the DUTS-TR~\cite{Zhao2015Saliency} dataset. MSRA10K includes 10,000 images with high quality pixel-wise annotations. This DUTS dataset is currently the largest saliency detection benchmark, and contains 10,553 training images (DUTS-TR) and 5,019 test images (DUTS-TE). Before feeding the training images into our proposed model, each image is rescaled into the same size [224,224] and normalized to [0,1] as well as the ground truth.

 After pre-training the master network, we can get the prior-model weight $\gamma$, which can guide the post-training weights $\theta$. Thus, we can obtain a prior-model guided saliency map $Sm^ \gamma$, denote as:
 \begin{equation}
Sm^ \gamma_{i,j}= p(S|R(I,i,j); \theta; \gamma),
\end{equation}
where $ \gamma$ is prior-model weight, which is fixed by the pre-training in master network.
\subsection{Depth-enhanced Subsidiary Network}

In order to obtain the features of an input depth map, we apply a subsidiary network to encode the depth cue and incorporate the depth-based features acquired by the subsidiary network as a convolution layer into the proposed master network. We denote the input depth map of this
convolution layer as $d$. Its corresponding output is:
 \begin{equation}
d_o= w \cdot d + b,
\end{equation}
where $b$ is the bias, and $w$ is the depth-enhanced weight matrix obtained by the subsidiary network.
\begin{figure*}
\begin{center}
\includegraphics[width=\textwidth]{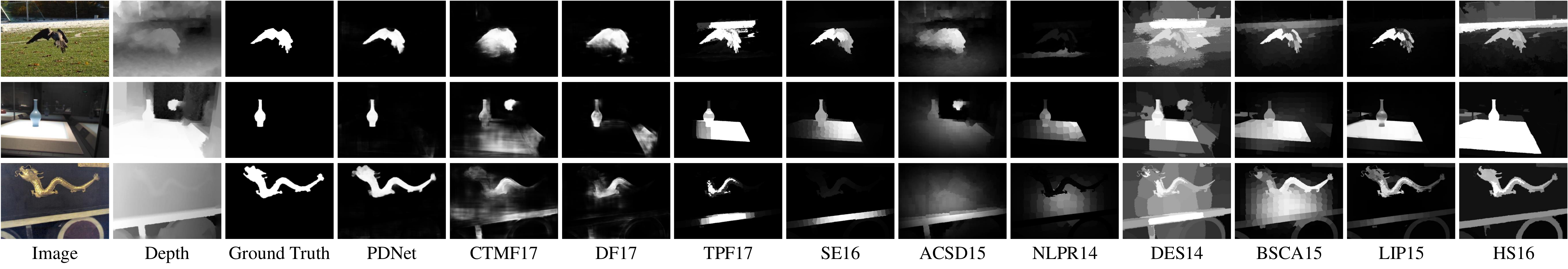}
\end{center}
\caption{The visual comparison of different methods.}
\label{fig:visual}
\end{figure*}
\begin{figure*}
\begin{center}
\includegraphics[width=\textwidth]{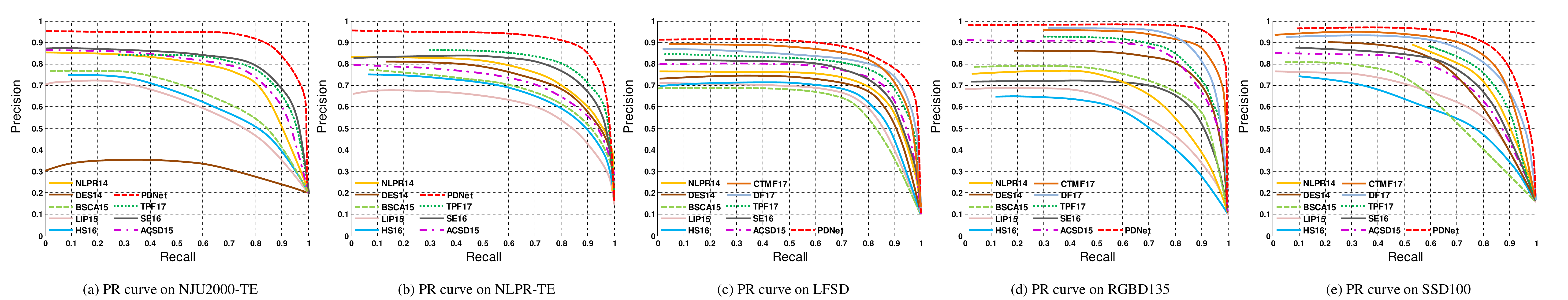}
\end{center}
\caption{The PR curve on five RGB-D datasets.}
\label{fig:pr}
\end{figure*}
\begin{figure}
\label{fig:4}
\begin{center}
\includegraphics[width=0.5\textwidth]{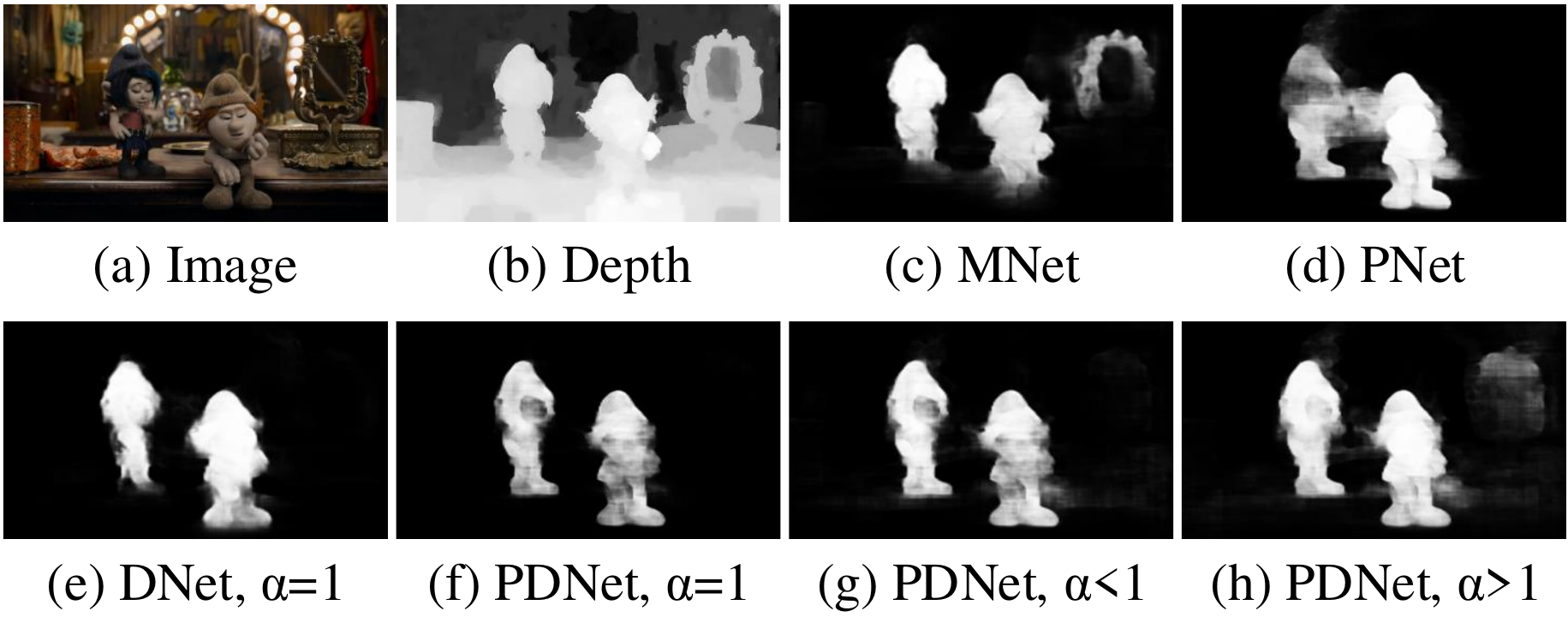}
\end{center}
\caption{The visual results of the ablation study.}
\label{fig:ablation}
\end{figure}

The output features $d_o$ of the subsidiary network is directly used as the weight matrix for the prior-model guided master network. The subsidiary network can therefore be viewed as a depth-enhanced weight prediction network to encode depth representation into the
master network. Eq. 3 can be rewritten as:
 \begin{equation}
Sm^{ \gamma, d_o}_{i,j}= p(S|R(I,i,j); \theta; \gamma; \alpha),
\end{equation}
where $\alpha$ is the combination weight factor of the depth-based feature maps obtained via sub-network, which is based on the number of feature maps, denote as:
 \begin{equation}
\alpha = \frac{d_o}{I_o},
\end{equation}
where $I_o$ is the RGB-based feature maps obtained via encoder part of prior-model guided master network.

\section{Experiments}

\subsection{Parameters and Running time}
The proposed model is implemented in python $2.7$ with Tensorflow $1.4$. It's evaluated on a machine equipped with an i7-7700 CPU and a Nvidia GTX1060 GPU (with 6G memory).
The parameters of hierarchical feature layers are initialized by the truncated normal distribution. The Adam optimizer is used for training with learning rate from 0.001 to 0.0001. The proposed master network of PDNet is pre-trained on large-scale (20,553) RGB datasets, which cost about 26 hours (15 epochs). Then we fix the parameters in encoder part of the proposed master network and train the proposed PDNet on RGB-D datasets. It takes about 10 minutes each epoch. While performing inference, it runs 15 fps (VGG-16) on average.

\subsection{Dataset}
In this section, we evaluate the proposed method on five RGB-D datasets.

{\bf NJU2000}~\cite{Ju2015Depth}. The NJUDS2000 dataset contains 2000 stereo images as well as the corresponding depth maps and manually labeled ground truths. The depth maps are generated using an optical flow method. We also randomly split this dataset into two parts: 1500 images for training and 500 for testing (NJU2000-TE).

{\bf NLPR}~\cite{Peng2014RGBD}. The NLPR RGB-D salient object detection dataset contains 1000 images captured by the Microsoft Kinect in
various indoor and outdoor scenarios. We randomly split this dataset into two parts: 500 images for training and 500 for testing (NLPR-TE).

{\bf LFSD}~\cite{Li2014Saliency}. This dataset contains 100 images with depth information and manually labeled ground truths. The depth information was captured using the Lytro light field camera. All images in this dataset were used for testing.

{\bf RGBD135}~\cite{Cheng2014Depth}. This dataset has 135 indoor images taken by Kinect with the resolution $640 \times 480$. All images in this dataset were used for testing.

{\bf SSD100}~\cite{8265566}. This dataset is built on three stereo movies. It contains 80 images with both indoors and outdoors scenes. All images in this dataset were used for testing.
\subsection{Evaluation Metrics}
Three most widely-used evaluation metrics used to evaluate the performance of different saliency algorithms, including the precision-recall (PR) curves, F-measure and mean absolute error (MAE).


\begin{table}
  \centering
\caption{Ablation study results on two RGB-D datasets.}
\label{ablation}
\begin{tabular}{|c|c|c|c|c|}
    \hline
    \multirow{2}{*}{Model}&
    \multicolumn{2}{c|}{NJU2000-TE~\cite{Ju2015Depth}} &
    \multicolumn{2}{c|}{ NLPR-TE~\cite{Peng2014RGBD}}\\ \cline{2-5}
    &${F}_{\beta}$ &MAE&${F}_{\beta}$&MAE\\
    \hline
    $MNet$ &  0.7610&0.0757 & 0.8193&0.0874 \\ \hline

    $PNet$ &  0.8397&0.0800 & 0.7901&0.0551 \\ \hline

    $DNet_{ \alpha = 1}$ & 0.8424     & 0.0728  & 0.8203     & 0.0630  \\  \hline

    $PDNet_{ \alpha = 1}$  & \textbf{0.8503}    & \textbf{0.0689}  & \textbf{0.8478}     & \textbf{0.0491} \\ \hline

    $PDNet_{ \alpha < 1}$  & 0.8471     &  0.0694 & 0.8409     & 0.0494     \\  \hline

    $PDNet_{ \alpha > 1}$   & 0.8458     & 0.0698 & 0.8284     & 0.0548 \\ \hline
\end{tabular}
\end{table}

\begin{table*}
  \centering
\caption{Quantitative comparison of F-measure and MAE scores on five RGB-D datasets. The best results are shown in bold font.}
\label{tab:compare}
\begin{tabular}{|c|c|c|c|c|c|c|c|c|c|c|}
    \hline
    \multirow{2}{*}{Method}&
    \multicolumn{2}{c|}{NJU2000-TE~\cite{Ju2015Depth}} & \multicolumn{2}{c|}{NLPR-TE~\cite{Peng2014RGBD}} & \multicolumn{2}{c|}{ LFSD~\cite{Li2014Saliency}} & \multicolumn{2}{c|}{ RGBD135~\cite{Cheng2014Depth}}& \multicolumn{2}{c|}{ SSD100~\cite{8265566}}\\ \cline{2-11}
    &${F}_{\beta}$ &MAE&${F}_{\beta}$&MAE&${F}_{\beta}$&MAE &${F}_{\beta}$&MAE&${F}_{\beta}$&MAE\\
    \hline
    PDNet & \textbf{0.8503}& \textbf{0.0689} & \textbf{0.8478}& \textbf{0.0491}      & \textbf{0.8219}& \textbf{0.0752}   & \textbf{0.8805}& \textbf{0.0384}  & \textbf{0.8152}& \textbf{0.0812} \\ \hline

    CTMF17~\cite{Han2017CNNs}   & -&- & -    & -  & 0.8025     & 0.0912      & 0.8102    &0.0653 & 0.7925     & 0.0912\\
    DF17~\cite{Li2017CNN}  & -&- & -     & -     & 0.8109     & 0.0815  & 0.8059     & 0.6871 & 0.7858     & 0.0846 \\
    TPF17~\cite{8265566}   & 0.7213&0.1488 & 0.7190&0.0852  & 0.7925     & 0.1058  & 0.7395     & 0.0891 & 0.7541     & 0.1217\\
    SE16~\cite{Guo2016Salient}  & 0.6946&0.1687 & 0.7101&0.0904 & 0.7568     & 0.1156  & 0.5807     & 0.1253 & 0.6666     & 0.1648\\
    ACSD15~\cite{Ju2015Depth}  & 0.6747&0.1939 & 0.6019&0.1624  & 0.7865     & 0.1425  & 0.6851     & 0.1518 & 0.6382     & 0.2010\\
    NLPR14\cite{Peng2014RGBD} & 0.6165&0.1669 & 0.5957&0.1087  & 0.7356     & 0.1547  & 0.4912     & 0.1165 & 0.6415     & 0.1784\\
    DES14~\cite{Cheng2014Depth} & 0.6202&0.4465 & 0.5915&0.3207  & 0.7254     & 0.2168  & 0.5410    & 0.3079 & 0.5797     & 0.3132\\ \hline

    BSCA15~\cite{qin2015saliency}  & 0.6290&0.2148 & 0.5925&0.1754  & 0.7126     & 0.1894  & 0.5826    & 0.1851 & 0.5755     & 0.2386\\
    LIP15~\cite{Li2015} & 0.5692&0.2059 & 0.5890&0.1252  & 0.7176     & 0.1880  & 0.5452     & 0.1406 & 0.5935     & 0.1960\\
    HS16~\cite{Shi2016} & 0.6090&0.2516 & 0.6003&0.1909  & 0.7248     & 0.1751  & 0.5361     & 0.1849 & 0.5716     & 0.2582\\
    \hline
\end{tabular}

\end{table*}

\subsection{Ablation Study}
To validate the effectiveness of our proposed network, we design a baseline and evaluate five variants of the baseline. The baseline is the master network of the proposed PDNet without prior-model guided and trained with four-dimensional RGB-D data ($MNet$). The five variants are:
\begin{itemize}
\item The master network of the proposed PDNet with prior-model guided and trained with three-dimensional RGB data ($PNet$).
\item The master network of the proposed PDNet without prior training, connecting the subsidiary depth-enhanced network ($DNet$, here we adopt $\alpha = 1$).
\item The proposed PDNet with the combination weight factor equals 1 ($PDNet_{ \alpha = 1}$).
\item The proposed PDNet with the combination weight factor less than 1 ($PDNet_{ \alpha < 1}$, here we take four samples, which are $\alpha = 0.3, 0.5, 0.7, 0.9$, and averaging them to present this situation).
\item The proposed PDNet with the combination weight factor larger than 1 ($PDNet_{ \alpha > 1}$, here we take four samples, which are $\alpha = 1.3, 1.5, 1.7, 1.9$, and averaging them to present this situation).
\end{itemize}

Table \ref{ablation} shows the MAE and F-measure
validation results on two RGB-D datasets. And the visual results of the ablation study is shown in Fig.\ref{fig:ablation}. We can clearly see the accumulated processing gains
after each component. In summary, it
proves that each variation in our algorithm is effective for generating the optimal final saliency map. And the approximate best performance is $PDNet_{ \alpha = 1}$, so we adopt $ \alpha = 1$ in the following experiments.
\subsection{Comparison with the State of the Art}

In this section, we compare our method with three state-of-the-art methods developed for
RGB images (BSCA15~\cite{qin2015saliency}, LIP15~\cite{Li2015}, and
HS16~\cite{Shi2016}) and seven RGB-D saliency methods designed specifically
for RGB-D images (DES14~\cite{Cheng2014Depth}, NLPR14\cite{Peng2014RGBD}, ACSD15~\cite{Ju2015Depth}, SE~\cite{Guo2016Salient}, TPF17~\cite{8265566}, DF17~\cite{Li2017CNN} and CTMF17~\cite{Han2017CNNs}). We use the codes provided by the authors to reproduce their experiments.
For all the compared methods, we use the default settings suggested by the authors.

Fig.\ref{fig:visual} provides a visual comparison of our approach with the above-mentioned approaches. It can be observed that our proposed method produce fine detail as highlighting the attention-grabbing salient region.

As shown in Table \ref{tab:compare}, our model outperforms other methods almost across all the datasets in terms of commonly-used evaluation metrics. From the PR Curve (Fig.\ref{fig:pr}) we can easily conclude that our approach achieves better results in all the five datasets.
\section{Conclusion}

In this paper, we propose a novel PDNet for RGB-D saliency detection. We adopt a prior-model guided master network to process RGB information of images. And the master network is pre-trained on the conventional RGB dataset to overcome the limited size of annotated RGB-D data. Instead of treating the depth map as a fourth-dimensional input, we design an independent sub-network for extracting depth information, which proves to be better than the previous treatment. Extensive experiments demonstrate that prior-model provides a solid foundation for salient object detection. And additionally integrating an independent depth-enhanced network contributes largely to the final accuracy. To encourage future works, we expose the source code that can be found on our project website: https://github.com/ChunbiaoZhu/PDNet/.

{\small
\bibliographystyle{IEEEbib}
\bibliography{bare_jrnl}
}

\end{document}